# A Deep Structure of Person Re-Identification using Multi-Level Gaussian Models

Dinesh Kumar Vishwakarma, *IEEE Member,* Sakshi Upadhyay

*Abstract*—Person re-identification is being widely used in the forensic, and security and surveillance system, but person re-identification is a challenging task in real life scenario. Hence, in this work, a new feature descriptor model has been proposed using a multilayer framework of Gaussian distribution model on pixel features, which include color moments, color space values and Schmid filter responses. An image of a person usually consists of distinct body regions, usually with differentiable clothing followed by local colors and texture patterns. Thus, the image is evaluated locally by dividing the image into overlapping regions. Each region is further fragmented into a set of local Gaussians on small patches. A global Gaussian encodes, these local Gaussians for each region creating a multi-level structure. Hence, the global picture of a person is described by local level information present in it, which is often ignored. Also, we have analyzed the efficiency of earlier metric learning methods on this descriptor. The performance of the descriptor is evaluated on four public available challenging datasets and the highest accuracy achieved on these datasets are compared with similar state-of-the-arts, which demonstrate the superior performance.

*Index Terms*—Feature Extraction, Feature Fusion, Image Matching, Image Recognition, Metric Learning, Person Re-Identification.

## I. INTRODUCTION

NOWADAYS, widespread networks of cameras are being used in various public places like railway stations, airports, hospitals, shopping malls, etc. which cover large areas and have non-overlapping viewpoints and thus provide a huge amount of relevant data. To utilize this data effectively for public safety applications we cannot rely on manual monitoring and thus need efficient automated systems, which can track person's activity across multiple cameras. Person re-identification (Re-Id) is an elementary task for this.

Person Re-Id aims to identify all the occurrences of a person of interest, in the data available through sensors (cameras), for a bounded region. It is mostly done by extracting some kind of visual property captured in the image. We assume that the data captured refers to a short duration of time so that the appearance of the person, specifically the clothes and the rare objects (like bag packs, purse, etc.), do not change. However, the difficulty level of this task is increased due to significant intra-class variations caused by the differences in illumination, pose and view angle of the camera. Moreover, due to inter-class similarity as a consequence of similar clothing and low resolution images acquired. This can be clearly seen in Fig.1 for some acquired public datasets. So the efforts are focused on developing discriminative features robust to these challenges and learning distance models for correct matching.

Most of the approaches followed in re-ID acquire appearance-based features [1] [2] [3] [4], such as color and texture models. To address the low resolution issue and varying poses, the color information like color histograms of desired color channels and color name descriptors [5] come to the rescue. However, they lack texture and spatial information and thus cannot sufficiently differentiate different persons of similar color. Thus texture descriptors, like Maximally Stable Color Regions (MSCR) [6], Local Binary Patterns (LBP) and 21 texture filters (8 Gabor filters and 13 Schmid filters), have been used in some works. However, it is seen earlier that texture descriptors do not give good results when used alone. Thus they are usually added as complimentary information to color and spatial features such that this fusion feature gives a rich quality signature. In [7], color histograms in two color spaces is combined with Gabor and Schmid filter responses and extracted on strips of predefined height and position. Maximally Stable Color Regions (MSCR) is fused with weighted Color Histograms (wHSV) in [8] [9], achieving outstanding results. Recently some high dimensional features, incorporating fusion, were also proposed. Some of them are SDALF [9], kBiCov

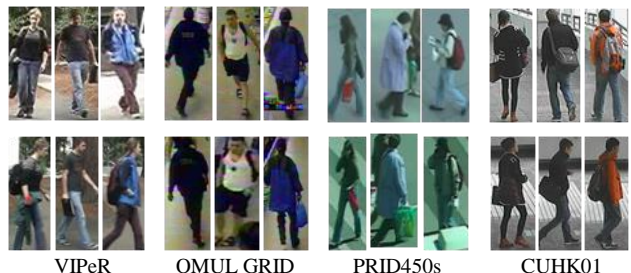

VIPeR     QMUL GRID     PRID450s     CUHK01
Fig. 1. Image pairs from VIPeR [4], QMUL GRID [29], PRID450s [30] and CUHK01 [31] datasets. Each column holds the images of the same individual captured from different cameras

[10], Fisher vectors (LDFV) [11], salience match [12], mid-level filter [13] and also ELF [4], where 8 color channels and 19 texture channels (Gabor and Schmid) were fused.

Methods for local feature encoding and pooling, as described in Fisher encoding [14] and Vector of locally Aggregated

D.K. Vishwakarma is working as an Associate Professor in the Department of Information Technology, Delhi Technological University, Delhi, India-11042 (e-mail: dinesh@dtu.ac.in, dvishwakarma@gmail.com)

S. Upadhyay is post graduate student in the Department of Electronics and Communication Engineering, Delhi Technological University, New Delhi, India-110042 (e-mail: sakshiupadhyay8@gmail.com).

Descriptors [15] (VLAD), are also the focus of many research works, since they have shown improvement on Bag of Words approach. One such popular feature with quite promising results is Covariance descriptor [16] [17] [18]. Earlier applied for matching and texture classification in [19] and then for pedestrian detection in [5], it estimated covariance of basic image features in overlapping regions. It gives a way to fuse multiple features naturally and also filters out noise samples during its computation, as it uses an average filter. To avoid overlapping of person a depth based video data [20] can be used but it results into a costly system and less field coverage in practical scenario. An image to video based person re-identification based on joint feature project matrix and heterogeneous dictionary pair learning [21] is developed to address the problem of variation within the videos.

In [21], the concept of shape of signal probability density function (SOSPDF) was presented which describes a signal based on its pdf, like a histogram and region covariance. However, an issue with using covariance is that it neglects the importance of mean information present in a local area. Therefore, it gave a descriptor named shape of Gaussians (SOG) as a new SOSPDF which inherited the properties of region covariance and also considered the mean information. Many other works also used such Gaussian models to improve the classification efficiency, like Huang and Wang [22] used it for face recognition. In [23] a Global Gaussian model was used to describe the probabilistic distribution of local features in an image. Some proposed to use Gaussian mixture model (GMM). In [24] hierarchical GMM was proposed in which the deviation of the distribution model of each image, with respect to the whole training set, was estimated instead of estimating GMM of each image, which would be impossible for a large scale data. However, it is not preferred due to its dependency on the training set. Another approach is Gaussian descriptor based on local features (GaLF) in which a 7-d feature including color, texture and spatial properties is extracted on each pixel and then the mean and covariance is calculated for region. However, it simply concatenates the two Gaussian parameters.

Inspired by these works and to overcome the drawbacks presented by them we have proposed a framework with multi-level encoding of pixel features, using Gaussian distribution. As we know that a human in an image has different body parts, with each part having some intra-similarity in features. So to utilize this local structure in a person image we have roughly divided the image into horizontal strips called regions. To create a deeper structure we have also worked on smaller consistent patches in the image with $k \times k$ pixel neighborhood. Then first we have estimated Gaussian for local patches and then for the entire region, by further extracting Gaussian parameters from the local patch Gaussians falling under that region. Former is called first level Gaussian (FG) and the latter is called second level Gaussian (SG). These region Gaussians are then concatenated to give the final descriptor. First level Gaussians are constructed on some kind of pixel features which can be color or texture data. For this, we have extracted color moments and Schmid filter responses in local patches, and Gradient information along with simple color values. Then we have created n-dimensional pixel features using different combinations of these basic pixel features and compared the performance.

Earlier a similar hierarchical model, covariance-of-covariance, has already been used effectively for image identification [25] in which region co-variance is approximated over local patch co-variances of pixel features. This is one of the sources of motivation for the use of two level distribution models in this work. However, the reason that it lacks the discriminative information provided by the mean data of pixels that's why we used the Gaussian model. Though, it handles the above issue but poses another challenge. Gaussian distribution lies on a particular Riemannian manifold, as per information geometry [26], where we cannot apply Euclidean functions while most existing discriminant metric methods just work in Euclidean space. Therefore, Riemannian manifold is projected onto a Euclidean space similar to the way described in [27] and [28], via. matrix logarithm and vectorization concept. Covariance-of-covariance approach [25] also used this type of solution to the Riemannian space problem.

In brief, the main contributions of the paper to the person re-identification are as follows:

- A deep structure is constructed using multilevel Gaussian model on four different combinations of colour, texture and spatial pixel features (colour moment and Schmid filter response).
- To make identification more effective a learning metric approach is applied on the developed structure and the identification accuracy is compared on four metric learning approaches.
- To evaluate the performance of the structure, an experiment is conducted on four challenging public datasets such as VIPeR [4], QMUL GRID [29], PRID450s [30] and CUHK01 [31]. The highest recognition rate is compared with the similar state-of-the-arts and demonstrate superior performance.
- In addition, to demonstrate the reliability of the structure, a person retrieval experiment is conducted and structure is able to re-identify the person.

The rest of paper is arranged as follows. The details of the proposed method is explained in section II, experimental and results are presented in section III and finally the conclusion of the work is given in section IV.

## II. MULTI-LEVEL GAUSSIAN DESCRIPTOR

We have constructed a deep structure of Gaussian model by extracting Gaussian distributions in two levels, one locally on the patch pixels, i.e. first level Gaussian and the other over the first level Gaussians in a defined region. The framework of our approach is shown in Fig 2. Motivation for this is taken from [25], [32], [33], [23] and [5]. It is named as a multi-level Gaussian descriptor. This structure is similar to some part based models described earlier where the image is divided into overlapping horizontal strips of fixed size given as one level of extraction and then each region is further dispersed into local square patches like the other level of extraction. Thus it addresses the non-rigid behavior of human body and captures the local information while representing the image globally.

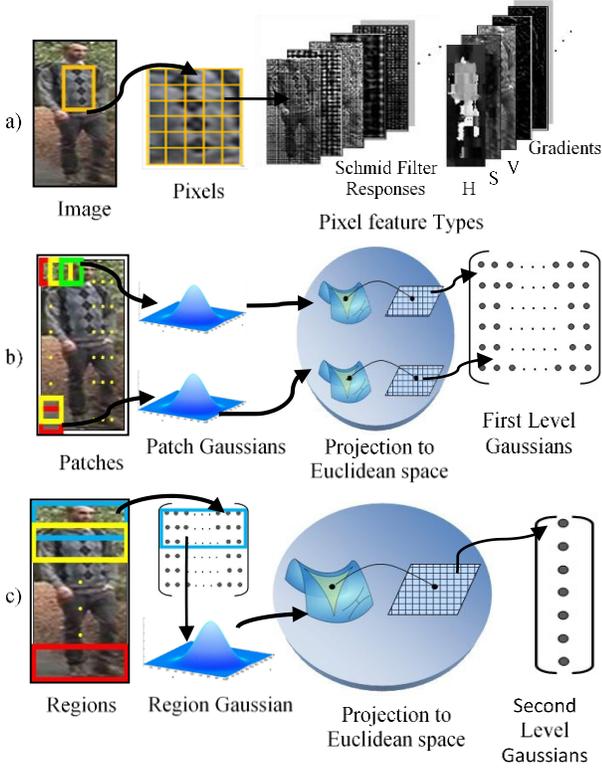

Fig. 2. Framework of the proposed Multi Level Gaussian Descriptor. It has three main sections (a) Extraction for low level pixel features (b) First level Gaussian encoding of patches (c) Second level Gaussian encoding of the regions over underlying patches.

The important aspect is to consider; the low-level pixel features, and it is necessary that the diverse aspects of the data contained in the image are extracted to add to the quality of base information on which further analysis is done. Thus in this method variety of base features are extracted and concatenated to create n-dimensional feature representation for each pixel before estimating the Gaussian model on them. The rest of the details are clearly given below.

*A. Low Level Features*

Define in this work, four different combinations of colour, texture and spatial pixel features are explored to create n-dimensional feature channel for each pixel in the image. Features used are color moment values of RGB components, HSV and normalized RGB color space values, Schmid filter responses, Gradient magnitudes in predefined orientations and spatial location of pixel.

Moments are statistical parameters that can uniquely describe a probabilistic distribution. Considering color in the image as a probabilistic distribution, moments on color values can provide a discriminative feature to measure the similarity between image pairs. They are inclusive of both shape and color data in the pixels and thus give an illumination, scale and rotation invariant feature, also used in [34] and [35]. In this work, first two moments of each R, G and B color values over local patches of $k \times k$ size, i.e. mean and standard deviation, are used in the feature channel. Color space values of $HSV$ color space and normalized $R$ and $G$ components are the other color based features used in the feature channel. Color values are normalized, as described in [36], to reduce the effect of a change in saturation levels, illumination and color response of the camera sensors. This is done by dividing the color channel value by the summation of all channel values of that pixel.

Then Schmid filter banks are used to capture texture and edge information. They have rotational invariance property and can be estimated by convolving isotropic "Gabor-like" filters [37] with the image. They are orientation sensitive, similar to Gabor, and give circular gradients in the image. These filters combine frequency and scale together as follows:

$$F(r, \tau, \sigma) = F_0(\tau, \sigma) + \cos(\frac{2\pi\tau r}{\sigma})\exp(-\frac{r^2}{2\sigma^2}) \quad (1)$$

Here, r depicts the radius, $\tau$ depicts the number of cycles of the harmonic function within the Gaussian envelope of the filters, $\sigma$ gives the scale and $F_0(\tau, \sigma)$ is a zero DC component which is added to obtain invariance to illumination changes. These filters are thus quite helpful in achieving invariance to intensity, pose and viewpoint.

Besides that, gradient information and spatial information of pixels are also added in the feature channel. Spatial location of a pixel in the vertical direction is used here because of the known symmetry present in the human body across the vertical axis, which is invariant to misalignments and viewpoint variations. The low-level features thus selected here have some useful properties to offer, but the selection is not too critical and finding better feature channels is very much a possibility.

*B. First Level Gaussian Encoding*

Once a low-level feature channel is extracted, image is transformed into a $W \times H \times n$ dimension representation where $n$ denotes the dimensions of feature channel $f_i$ for each pixel $i$ and, $W \times H$ gives the image size. These pixel features are then encoded by a Gaussian distribution calculated over small patches of k x k pixels with u step size. For each patch P, the Gaussian model $\mathcal{N}(f; \mu_p, \Sigma_p)$ is defined as:

$$\mathcal{N}(f; \mu_p, \Sigma_p) = \frac{\exp(-\frac{1}{2}(f-\mu_p)^T \Sigma_p^{-1}(f-\mu_p))}{(2\pi)^{n/2}|\Sigma_p|} \quad (2)$$

Here $\mu_p$ is the mean vector and $\Sigma_p$ is the covariance matrix estimated as $\mu_p = \frac{1}{N}\sum_{i \in \mathcal{A}_p} f_i$ and $\Sigma_p = \frac{1}{N-1}\sum_{i \in \mathcal{A}_p}(f_i - \mu_p)(f_i - \mu_p)^T$, respectively. $\mathcal{A}_p$ denotes the area of sampled patch P and N gives the number of pixels in $\mathcal{A}_p$. $f, \mu_p \in \mathbb{R}^n$ and $\Sigma_p \in \mathbb{S}_n^+$ where $\mathbb{S}_n^+$ is the space of real symmetric positive semi-definite (SPD) matrices. Estimated covariance matrix is $n \times n$ dimensional with diagonal entries representing the variance of each feature, and the non-diagonal entries giving their respective correlations.

So, in this step, the first level of Gaussian encoding is extracted over the pixel features. But for the second level of Gaussian encoding, we need to evaluate two mathematical parameters, i.e. mean and covariance, on a set of first level Gaussians. As per information geometry [26], Gaussians belong to Riemannian manifold where Euclidean functions cannot be applied. So it needs to be projected from the Riemannian space to a tangent Euclidean space favored by Riemannian metric, as described in [5] [38] [39] [27]. This mapping is performed over a Symmetric positive definite matrix. So firstly, the n-dimensional Gaussian distributions are embedded into a n+1 dimensional SPD matrix space $\mathbb{S}_{n+1}^+$, similar to the work in [28], as follows:

$$\mathcal{N}(f; \mu_p, \Sigma_p) \sim S_p = |\Sigma_p|^{-1/(n+1)} \begin{bmatrix} \Sigma_p + \mu_p \mu_p^T & \mu_p \\ \mu_p^T & 1 \end{bmatrix}, \quad (3)$$

where $S_p$ is the $(n+1) \times (n+1)$ dimensional SPD matrix derived from the local Gaussians. This is then mapped onto a space tangent to the manifold, at the tangency point given by matrix T, via simple matrix logarithm, as defined in [5]:

$$s = log_T(S_p) \triangleq T^{\frac{1}{2}} \log(T^{-\frac{1}{2}} S_p T^{-\frac{1}{2}}) T^{\frac{1}{2}} \quad (4)$$

Here, s is the tangent vector in the tangent space at point T, which is further minimized by vectorization, given as:

$$S_p' = vec_T(s) = vec_I(T^{-\frac{1}{2}} s T^{-\frac{1}{2}}) \quad (5)$$

$S_p'$ denotes the orthonormal coordinates of $S$ which lie in a Euclidean space and $I$ denotes an identity matrix. Replacing the value of '$S$' in (5) by the value defined in (4) redefines $S_p'$, the projection vector of $S_p$, as follows:

$$S_p' = vec_I(\log(T^{-\frac{1}{2}} S_p T^{-\frac{1}{2}})) \quad (6)$$

Therefore, the matrix of patch Gaussian $S_p$ becomes $m = \frac{(n^2+3n)}{2} + 1$ dimensional vector $S_p'$.

### C. Second Level Gaussian Encoding

Now the second level of encryption is done over the patch Gaussians obtained previously. Each image is roughly divided into seven overlapping horizontal stripes, called regions, to utilize the local geometry in humans. Then these regions are again encoded by a Gaussian distribution $\mathcal{N}(S'; \mu^R, \Sigma^R)$ estimated over the first level Gaussians falling under the region R. Thus a discriminative and compact model of Gaussian distributions is learnt. But to use this feature for classification and matching purpose, again we need to apply SPD matrix conversion and projection to Euclidean space, as we did previously. This is required because most of the existing metric methods work in Euclidean space. Then the SPD matrix obtained will have $(m+1) \times (m+1)$ dimensions and the projected vector $H$ will have $(m^2 + 3m/2) + 1$ dimensions.

The final feature vector is then formed by just concatenating all the region feature vectors $\{H_r\}_{r=1}^R$ into a single feature vector $H = [H_1^T, \ldots, H_R^T]^T$. Due to deep structure of the model final feature vector is very high dimensional and thus L2 norm normalization is performed. But prior to this, mean removal is done on the feature vectors to deal with the large differences in the values at different dimensions, as described in [40]. This may occur due to the use of variety of pixel features with different distribution characteristics. Then the normalization is defined as: $H = (H - \bar{H})/\| H - \bar{H} \|_2$ where $\bar{H}$ is the mean vector of the training set feature vectors.

### D. Metric Learning Methods

Once the descriptor is constructed then, we need to compare the derived features to find the most similar resemblances. For this, robust and efficient distance metrics must be used. They must try to minimize the distance values between the same class objects while maximizing those from different classes. In this work, four metric learning methods are compared against the proposed descriptor. These are SVMML [41], LFDA [42], KISSME [43] and XQDA [32].

SVMML, proposed in [41], learns a decision function locally by using a distance metric and a locally adaptive threshold rule jointly, following SVM-like objective function approach. LFDA proposed in [42] learns a distance metric using a two stage processing, where first PCA is applied for dimensionality reduction and then supervised FDA is used along with LPP for further reduction. Then KISSME was proposed in [43] as an effective metric just based on equivalence constraints. It does not involve tedious iterative computation for optimizing the procedure. It just needs to compute two small sized co-variance matrices and is thus scalable to large datasets. An extension of KISSME, LFDA and Bayesian Metric methods was then proposed in LOMO [32], named as Cross view discriminant analysis or XQDA. The statistical inference perspective based on likelihood-ratio test as given in KISSME also inspires it. It can learn a lower dimension discriminant subspace and a metric kernel on it, simultaneously, by following generalised Rayleigh-quotient formulation. While LFDA, after deriving a subspace directly uses the Euclidean function on it, XQDA further approach to use an efficient metric as well.

## III. EXPERIMENTS AND RESULTS

To show the efficiency of our method we have performed various experimental studies, including comparisons with state-of-the-art descriptors and metrics, on four challenging datasets: VIPeR [4], QMUL GRID [29], PRID450s [30] and CUHK01 [31]. The results are presented in the form of Cumulative Matching Characteristics (CMC) curves and Recognition rates at different ranks. We have also performed image retrieval experiment on some of the datasets considering person re-identification as a recognition problem.

### A. Datasets and Setup

*VIPeR* [4] has a total of 1264 images of 632 individuals captured from two different camera views, A and B. The images have illumination and viewpoint variations, and are cropped and resized to $128 \times 48$ pixels. For evaluation, the widely adopted protocol is used where the set of 632 image pairs is randomly split into training, and a testing set with 316 image pairs in each and 10 such sets are created to produce average results.

*QMUL GRID* [29] contains 250 low-resolution pedestrian image pairs captured in a busy underground station by two cameras and contains variation in poses and lighting. Two folders are provided namely 'Probe' and 'Gallery' each containing 250 images of 250 individuals while gallery folder also contains additional 775 images that do not belong to probe set and are used in the testing set during cross validation.

*PRID450S* [30] contains 450 images captured from disjoint views by two cameras A and B. For evaluation, 225 individuals are assigned randomly to training and testing set.

*CUHK01* [31] has 3884 images in a total of 971 individuals manually cropped to 60x160 pixels. Two disjoint camera view were selected where camera A captures more pose and viewpoint variations while camera B captures mainly frontal view and back view. Each person has two images captured in each camera. The training set contains 486 while testing set contains 485 individuals selected randomly.

Each image in all these datasets is resized to 128×48 pixels. For each image pair, image from one camera is assigned as a probe and the other camera as a gallery, randomly. It is done in both training and testing phase. Then the process of selecting a single image from the probe set and matching it with all images from the gallery set is repeated for all images in the probe set.

### B. Feature settings

We have used four different sets of n-dimensional pixel level feature vector.

1) *Pixel Feature 1 (YCM)*

In this y distance and first two color moments, i.e. mean and standard deviation of RGB color space are used to create a 7-d pixel feature. Color moments are calculated on patches of size $5 \times 5$ with patch interval of size 2. It can be represented as:

$[y, mean(R), std(R), mean(G), std(G), mean(B), std(B)]$

2) *Pixel Feature 2 (SCHMID)*

This is a texture feature representation of pixels comprising of 13-d schmid filter responses using schmid filter banks applied on $10 \times 10$ non-overlapping patches. Parameter values of $(\sigma, \tau)$ for 13 filter banks are (2,1), (4,1), (4,2), (6,1), (6,2), (6,3), (8,1), (8,2), (8,3), (10,1), (10,2), (10,3) and (10,4).

3) *Pixel Feature 3 (YGOHSV*

The third kind of feature combination used has 1-d y distance value, 4-d gradient orientation and 3-d HSV color space values, i.e. a 8-d pixel feature representation.

4) *Pixel Feature 4 (YGOnRnG)*

It is a 7-d pixel feature representation comprising of 1-d y distance value, 4-d gradient orientation and 2-d nRnG color space values. Here the nRGB is the normalized color space obtained by normalising RGB color space values (e.g., nR = R/(R+G+B)). We have used only {nR, nG} values, since this color space has redundancy.

After constructing n-dimensional pixel feature we have extracted multi-level Gaussian descriptor for seven horizontal strips or regions (R=7) that are overlapping. Each strip comprises $32 \times 48$ pixels. In each region local patch Gaussians are extracted from the patches of $5 \times 5$ pixels with patch interval 2.

### C. Performance Comparison of Different Pixel Features

The effectiveness of each pixel feature and their fusion is evaluated separately, and the results are shown by CMC curves in Fig. 3, Fig. 4 and Fig. 5, for VIPeR, PRID450s and QMUL GRID, respectively. The identification rates at rank-1, rank-10

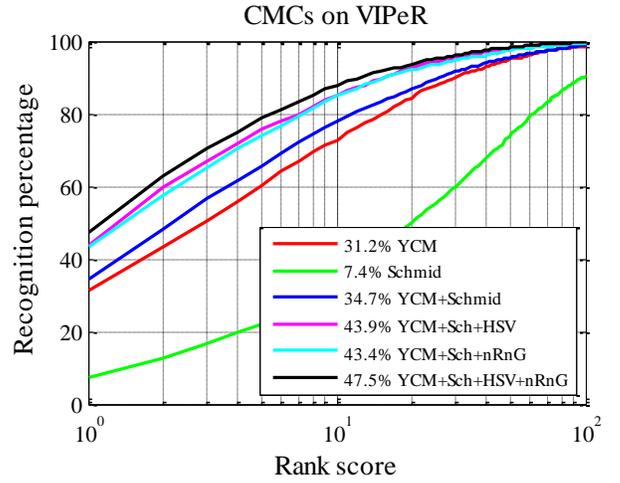

Fig. 3. CMC curve and Rank-1 identification rates on VIPeR

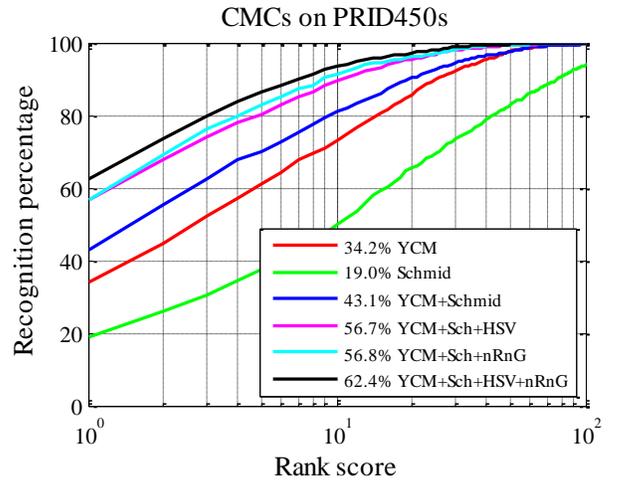

Fig. 4. CMC curve and Rank-1 identification rates on PRID450s

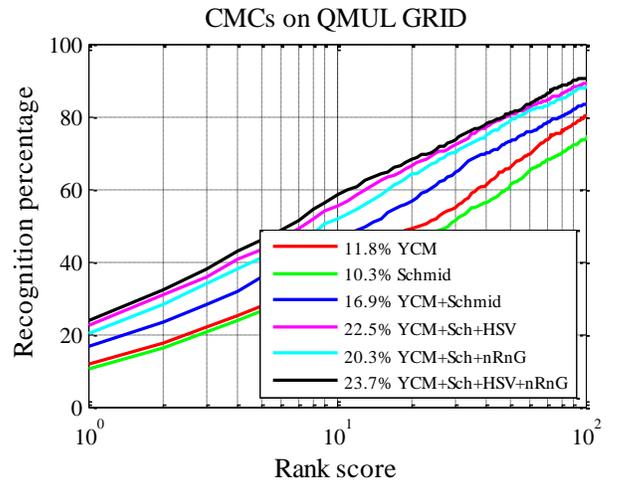

Fig. 5. CMC curve and Rank-1 identification rates on PRID450s

and rank-20, for different combinations of pixel features, are given in Table I. XQDA metric is used for this evaluation. Rank-1 recognition rate of YCM pixel feature is highly greater than the pixel feature comprising of Schmid filter responses, for VIPeR and PRID450s. It proves that for person re-id problem,

TABLE I
COMPARISON OF RECOGNITION RATES FOR DIFFERENT COMBINATION OF PIXEL FEATURES ON VIPER, PRID450S AND GRID DATASETS. THE BEST RESULTS ARE HIGHLIGHTED IN BOLD.

| Dataset | Pixel Feature Used | R=1 (%) | R=10 (%) | R=20 (%) |
|---|---|---|---|---|
| VIPeR | YCM | 31.2 | 72.9 | 86.8 |
|  | Schmiid | 7.4 | 34.2 | 50.1 |
|  | YCM+Schmid | 34.7 | 78.1 | 86.8 |
|  | YCM+Schmid+HSV | 43.9 | 85.1 | 92.5 |
|  | YCM+Schmid+nRnG | 43.4 | 85.1 | 92.0 |
|  | YCM+Schmid+HSV+nRnG | **47.5** | **87.9** | **93.7** |
| PRID450s | YCM | 34.2 | 73.3 | 85.7 |
|  | Schmiid | 19.0 | 49.8 | 65.3 |
|  | YCM+Schmid | 43.1 | 81.1 | 90.3 |
|  | YCM+Schmid+HSV | 56.7 | 89.6 | 95.4 |
|  | YCM+Schmid+nRnG | 56.8 | 91.2 | 96.0 |
|  | YCM+Schmid+HSV+nRnG | **62.4** | **93.5** | **96.9** |
| QMUL GRID | YCM | 11.8 | 38.2 | 49.2 |
|  | Schmiid | 10.3 | 35.4 | 49.2 |
|  | YCM+Schmid | 16.9 | 45.7 | 56.7 |
|  | YCM+Schmid+HSV | 22.5 | 55.3 | 66.4 |
|  | YCM+Schmid+nRnG | 20.3 | 52.0 | 64.2 |
|  | YCM+Schmid+HSV+nRnG | **23.7** | **58.2** | **68.1** |

*YCM+Schmid+HSV+nRnG = proposed descriptor (MLGD).
Evaluation performed using XQDA metric.

color features perform much better than texture features. But texture feature when combined with color feature, exceeds the rank-1 recognition rates of those using the color feature alone. The best results are achieved when all the four feature types are fused to make a rich signature. We have proposed this fusion descriptor as Multi-Level Gaussian descriptor (MLGD).

### D. Performance Comparison of the Proposed Descriptor on Different Metrics

The proposed descriptor MLGD is evaluated by four different metric learning methods namely, SVMML [41], LFDA [42], KISSME [43] and XQDA [32]. The results on VIPeR dataset are given in Fig. 6 by CMC curves and the recognition rates at rank-1, 10 and 20 are given in Table II. It is clearly visible that the proposed descriptor gives good rank-1 recognition rates, greater than > 35%, when evaluated with these state-of-the-art metric learning methods. Thus it can be inferred that the proposed descriptor holds a robust structure and is an effective representation for images. Best results, with rank-1 recognition rates of 47.53%, are achieved by using XQDA metric.

### E. Performance Comparison with State-of-the-Art Methods

The proposed descriptor is compared with both the single as well multi-level descriptors and the performance is listed in Table III. Analysis on covariance distribution models show that the Covariance descriptor when used globally on the image as proposed in [19], gives 26.9% rank-1 rates on VIPeR dataset while a multi-level model of covariance, proposed as Cov-of-Cov in [44] [25], gives 33.9% rank-1 rates on VIPeR. It is approximately 7% better than the global model of covariance.

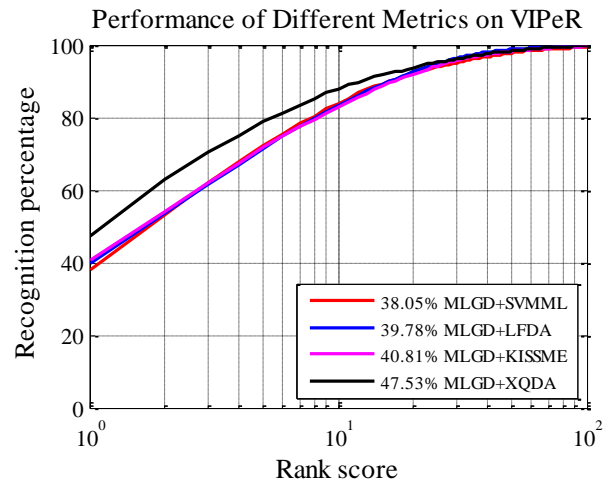

Fig. 6. CMC curve and Rank-1 identification rates of the descriptor evaluated on four different metrics.

TABLE II
COMPARISON OF RECOGNITION RATES FOR STATE-OF-THE-ART METRICS APPLIED ON THE PROPOSED DESCRIPTOR. BEST RESULTS ARE HIGHLIGHTED IN BOLD

| Metrtic Methods | Rank=1 (%) | Rank=10 (%) | Rank=20 (%) |
|---|---|---|---|
| MLGD+SVMML [41] | 38.05 | 83.90 | 92.02 |
| MLGD+LFDA [42] | 39.78 | 83.21 | 92.59 |
| MLGD+KISSME [43] | 40.81 | 82.80 | 91.85 |
| MLGD+XQDA [32] | **47.53** | **87.92** | **93.71** |

There are number of single-level meta descriptors such as Heterogeneous Auto-Similarities of Characteristics (HASC) [45], Hybrid Spatiogram and Covariance Descriptor (HSCD) [16], Local Descriptors encoded by Fisher Vector (LDFV) [11], Second-order Average Pooling (2AvgP) [46], and GOLD [47] are also chosen for comparison of the performance listed in Table III. They are non-hierarchical models and ignore the local properties of regions. They have similar performances. It can be seen that our descriptor has the highest recognition rates, with the rank-1 rate of 47.5% on VIPeR. It is also evident from the table that both the hierarchical methods, i.e. our method and Cov-of-Cov [25], have higher rank-1 matching rates than the non-hierarchical ones and thus it can be inferred that hierarchical model usually performs better. Between the two hierarchical methods mentioned, the difference is that we have used mean information together with the covariance value while [25] has used just the covariance information. Since our method has 13.6 % better rank-1 rates than hierarchical covariance descriptors so this proves that means information is a valuable feature of the image.

In section (c) of Table III, we have listed the reported results on some of the state-of-the-art method, including Metric Ensemble [48], Mid-Level Filter Learning (MLFL) [13], Salience Matching [12] and LOMO [32]. It can be clearly noted from the results that the performance of MLGD goes above many state-of-the-art methods with 47.5%, 62.4%, 54.5% and 23.7% rank-1 rates on VIPeR, PRID450S, CUHK01 and GRID dataset, respectively. The proposed descriptor even outperforms the efficient metric ensemble [48] by 1.6 % rank-1 rates.

TABLE III.
PERFORMANCE COMPARISON OF THE PROPOSED DESCRIPTOR MLGD AND STATE-OF-THE-ART DESCRIPTORS ON VIPeR, PRID450S, QMUL GRID AND CUHK01. SECTION (A) AND (B) REPRESENTS MULTI LEVEL DESCRIPTORS AND SINGLE LEVEL DESCRIPTORS, RESPECTIVELY, EVALUATED USING XQDA METRIC. SECTION (C) SHOWS THE PERFORMANCE OF SOME STATE-OF-THE-ART METHODS. THE BEST RESULTS IN EACH SECTION ARE GIVEN IN RED AND THE SECOND BEST IN BLUE.

| Section | Descriptors | VIPeR | | | PRID450s | | | GRID | | | CUHK01 | | |
|---|---|---|---|---|---|---|---|---|---|---|---|---|---|
| | | R=1 (%) | R=10 (%) | R=20 (%) | R=1 (%) | R=10 (%) | R=20 (%) | R=1 (%) | R=10 (%) | R=20 (%) | R=1 (%) | R=10 (%) | R=20 (%) |
| (A) | MLGD(Our method) | 47.5 | 87.9 | 93.7 | 62.4 | 93.5 | 96.9 | 23.7 | 58.2 | 68.1 | 54.5 | 83.5 | 90.5 |
| | COV-of-COV [25] | 33.9 | 76.6 | 87.7 | 47.0 | 83.4 | 91.6 | 16.6 | 45.0 | 55.2 | 40.9 | 72.5 | 81.1 |
| (B) | Cov [19] | 26.9 | 65.8 | 77.1 | 40.4 | 73.4 | 82.1 | 10.6 | 29.0 | 36.7 | 34.5 | 64.5 | 73.6 |
| | GOLD [47] | 27.1 | 66.5 | 77.7 | 40.5 | 73.8 | 82.2 | 10.9 | 29.2 | 37.4 | 35.3 | 65.2 | 74.2 |
| | HSCD [16] | 31.2 | 86.5 | 91.8 | - | - | - | - | - | - | - | - | - |
| | 2AvgP [46] | 28.8 | 68.5 | 79.2 | 44.7 | 75.8 | 83.8 | 12.9 | 36.7 | 47.4 | 36.1 | 68.1 | 76.3 |
| | LDFV [11] | 25.3 | 66.8 | 79.4 | 32.1 | 66.9 | 77.6 | 16.2 | 41.9 | 53.1 | 36.4 | 71.0 | 80.3 |
| | HASC [45] | 30.9 | 70.6 | 81.8 | 41.8 | 76.3 | 85.2 | 12.9 | 35.6 | 47.3 | 38.6 | 68.7 | 77.1 |
| | CH+LBP [49] | 27.7 | 69.3 | 82.4 | 21.5 | 60.8 | 74.4 | 16.2 | 45.0 | 57.1 | 31.3 | 70.4 | 81.5 |
| (C) | Metric Ensemble [48] | 45.9 | 88.9 | 95.8 | - | - | - | - | - | - | 53.4 | 84.4 | 90.5 |
| | LOMO [32] | 41.1 | 82.2 | 91.1 | 62.6 | 92.0 | 96.6 | 17.9 | 46.3 | 56.2 | 49.2 | 84.2 | 90.8 |
| | SalMatch [12] | 30.2 | 65.0 | - | - | - | - | - | - | - | 28.5 | 55.0 | - |
| | MLFL [13] | 29.1 | 65.9 | 70.9 | - | - | - | - | - | - | 34.3 | 65.0 | 75.2 |

*F. Image Retrieval Results*

We have also performed an image retrieval experiment where we have tried to retrieve the true match of a probe image from the disjoint camera images. We have performed this on three datasets namely, VIPeR, CUHK01 and PRID450s and displayed the results in the Fig. 7, Fig. 8 and Fig. 9.

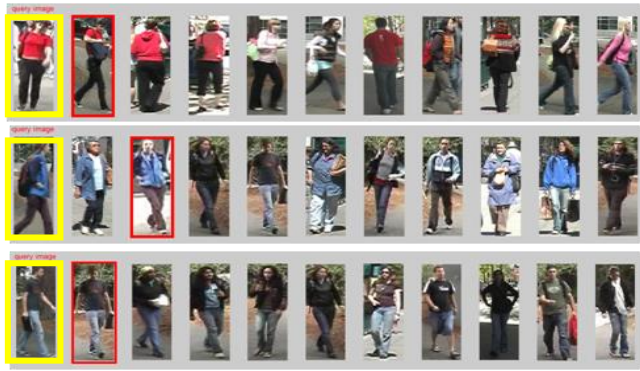

Fig. 7. Image retrieval results on VIPeR. Yellow boxes represent the query image and Red boxes give the true match pair of the query image.

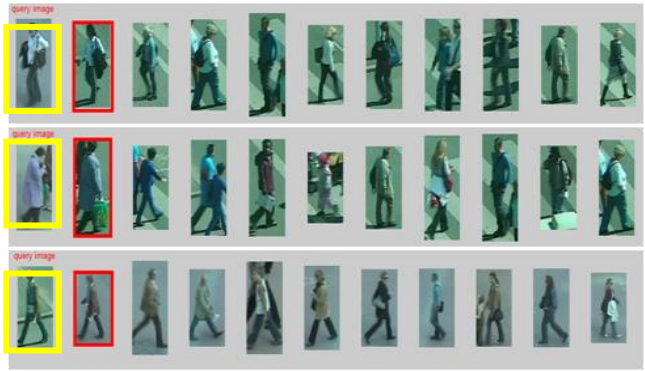

Fig. 8. Image retrieval results on PRID450s Yellow boxes represent the query image and Red boxes give the true match pair of the query image.

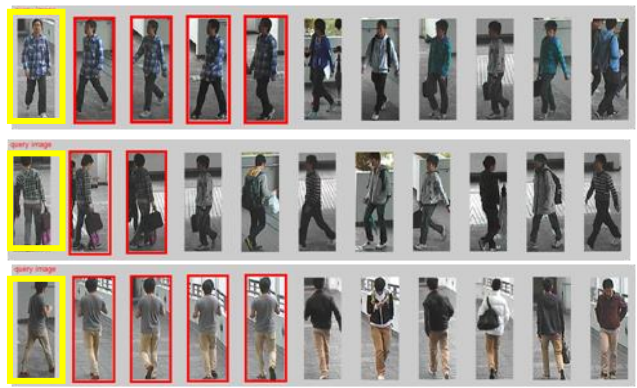

Fig. 9. Image retrieval results on CUHK01. Yellow boxes represent the query image and Red boxes give the true match pair of the query image.

We have first given a probe image to the system and then retrieved closest 10 images from the disjoint gallery set. Out of these, one of the 10$^{th}$ retrieved images contains the true match of the probe image from a different view angle and is highlighted by Red boxes in the given figures. The results clearly show that the recognition accuracy of the system is quite high, as the closest matches found are also the true match pair of the probe image. It must be mentioned here that while VIPeR and PRID450s have single frame per person in each camera images, CUHK01 has two or more image frames per person in each camera images.

IV. CONCLUSION

This work presents an efficient and effective approach of multi-layer Gaussian descriptor model for the problem of person re-ID. The descriptor presented in this work utilizes both, mean and covariance values of pixel features, present in an image and thus returns a robust and discriminative representation of data. It first encodes the local patches by a Gaussian distribution and then performs a second level of encoding over a set of local patches falling under overlapping regions, again by a Gaussian model. In this way it does not neglect the relevant information in the local structures of the image while acquiring a global representation. The results of our in depth experiments proved that the proposed descriptor can even outperform the state-of-the-art performances on four public datasets. We have also conducted image retrieval experiment where we treat this problem as a recognition problem. The results prove that our descriptor is robust against viewpoint changes and illumination variations. Further we have analyzed four different efficient metrics on our descriptor to show its robust structure.

In future the deep structure of Gaussian can be explored further to consider the local structure of human appearances in more depth. There is ample of scope in finding computationally low cost features to reduce the extraction and search time of the system.

In addition, some other kinds of pixel features can be tested to search for any further improvements in the process accuracies since the selection of these features is not too critical and finding better feature channels is very much a possibility. Then we can also try to learn a new metric, which is highly discriminative for high dimensional feature vectors.

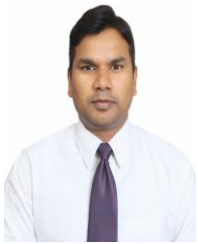
**Dinesh Kumar Vishwakarma** received the B.Tech. degree from Dr. Ram Manohar Lohia Avadh University, Faizabad, India, in 2002, the M.Tech. degree from the Motilal Nehru National Institute of Technology, Allahabad, India, in 2005, and the Ph.D. degree in computer vision from Delhi Technological University, New Delhi, India, in 2016. He is currently an Associate Professor with the Department of Information Technology, Delhi Technological University, Delhi, India. His current research interests include human action and activity recognition, hand gesture recognition, gait analysis, and machine learning.

Dr. Vishwakarma is a Reviewer of various journals of IEEE/IET, Springer, and Elsevier.

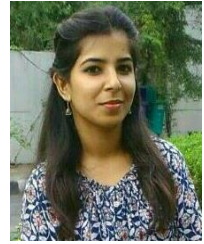
**Sakshi Upadhyay** Received B.Tech degree in Electronics and Communication Engineering from BT Kumaon Institute of Technology, Dwarahat, Uttarakhand, India in 2014, and M.Tech. in Signal Processing and Digital Design from Delhi Technological University, New Delhi, in 2017. Her research interests include Person Identification, Digital Image Processing, Machine learning and Pattern Recognition.